\documentclass[letterpaper, 10 pt, journal, twoside]{IEEEtran}

\IEEEoverridecommandlockouts                              

\usepackage{cite}
\usepackage{amsmath,amssymb,amsfonts}
\usepackage{algorithmic}
\usepackage{graphicx}
\usepackage{textcomp}
\usepackage{xcolor}
\usepackage{bbold}
\usepackage{url}
\usepackage{bbm}
\usepackage{subcaption} 
\usepackage{wrapfig}
\newcommand{\Contributions}[2]{ {\setlength{\parindent}{\paperwidth} \textbf{ \noindent C{#1}}\quad\emph{#2} } \newline}

\newcommand{\dmp}[1]{{\textcolor[rgb]{0.51, 0.45, 0.70}{#1}}}
\newcommand{\coupling}[1]{{\textcolor[rgb]{0.39, 0.71, 0.8}{#1}}}
\newcommand{\forcing}[1]{{\textcolor[rgb]{0.87, 0.53, 0.34}{#1}}}
\newcommand{\task}[1]{{\textcolor[rgb]{0.33, 0.67, 0.41}{#1}}}

\newcommand{\ra}[1]{{\textcolor[rgb]{0.0,0.0,0.0}{#1}}}
\newcommand{\rano}[1]{{\textcolor[rgb]{0.0,0.0,0.0}{#1}}}
\newcommand{\rb}[1]{{\textcolor[rgb]{0.0,0.0,0.0}{#1}}}
\newcommand{\rbno}[1]{{\textcolor[rgb]{0.0,0.0,0.0}{#1}}}

\begin{document}

\title{Residual Learning from Demonstration: \\ Adapting DMPs for Contact-rich Manipulation}

\author{Todor Davchev$^{\dagger \bullet \ast}$ Kevin Sebastian Luck$^{\odot}$ Michael Burke$^{\circ}$ Franziska Meier$^{\diamond}$ \\ Stefan Schaal$^{\ddagger}$ and Subramanian Ramamoorthy$^{\dagger}$
\thanks{Manuscript received: September, 9, 2021; Revised: December, 4, 2021; Accepted: January, 24, 2022.}
\thanks{This paper was recommended for publication by Editor Tetsuya Ogata upon evaluation of the Associate Editor and Reviewers' comments.}%
\thanks{This research was supported by: the EPSRC iCASE award with Thales Maritime Systems provided to Todor Davchev; %
EPSRC UK RAI Hub NCNR (EPR02572X/1) and Academy of Finland Flagship programme: Finnish
Center for Artificial Intelligence funding Kevin S. Luck; %
the Alan Turing Institute, as part of the Safe AI for surgical assistance project, funding Subramanian Ramamoorthy and Michael Burke at the University of Edinburgh.
}%
\thanks{$^{\ast}$ Corresponding Author.}%
\thanks{$^{\bullet}$ Research done in part during a Google X AI Residency.}%
\thanks{$^{\dagger}$ Authors are with the School of Informatics, University of Edinburgh, 10 Crichton St, EH8 9AB, United Kingdom, {\tt\small \{t.b.davchev, s.ramamoorthy\}@ed.ac.uk}}%
\thanks{$^{\odot}$ Author is with the Department of Electrical Engineering and Automation, Intelligent Robotics, Aalto University, 02150 Espoo, Finland, {\tt\small kevin.s.luck@aalto.fi
}}%
\thanks{$^{\circ}$ Author is with ECSE, Monash University, Melbourne, Australia, {\tt\small michael.g.burke@monash.edu}}
\thanks{$^{\ddagger}$ Author is with [Google X] Intrinsic, Mountain View, CA, USA, {\tt\small stefan.k.schaal@gmail.com}}%
\thanks{$^{\diamond}$ Author is with Facebook AI Research, Menlo Park, CA
{\tt\small fmeier@fb.com}}%
\thanks{Digital Object Identifier (DOI): see top of this page.}
}

\markboth{IEEE Robotics and Automation Letters. Preprint Version. Accepted January, 2022}{Davchev \MakeLowercase{\textit{et al.}}: Residual Learning from Demonstration}

\maketitle

\begin{abstract}

Manipulation skills involving contact and friction are inherent to many robotics tasks. Using the class of motor primitives for peg-in-hole like insertions, we study how robots can learn such skills. Dynamic Movement Primitives (DMP) are a popular way of extracting such policies through behaviour cloning (BC) but can struggle in the context of insertion. Policy adaptation strategies such as residual learning can help improve the overall performance of policies in the context of contact-rich manipulation. However, it is not clear how to best do this with DMPs. As a result, we consider several possible ways for adapting a DMP formulation and propose ``residual Learning from Demonstration`` (rLfD), a framework that combines DMPs with Reinforcement Learning (RL) to learn a residual correction policy. Our evaluations suggest that applying residual learning directly in task space and operating on the full pose of the robot can significantly improve the overall performance of DMPs. We show that rLfD offers a gentle to the joints solution that improves the task success and generalisation of DMPs \rb{and enables transfer to different geometries and frictions through few-shot task adaptation}. The proposed framework is evaluated on a set of tasks. A simulated robot and a physical robot have to successfully insert pegs, gears and plugs into their respective sockets. Other material and videos accompanying this paper are provided at \textit{https://sites.google.com/view/rlfd/}.

\end{abstract}

\section{INTRODUCTION}
\label{sec:introduction}
\IEEEPARstart{P}{art} insertion, e.g. plugs, USB connectors, house keys or car refuelling nozzles, is a manipulation skill required in a variety of practical applications, ranging from the home to manufacturing environments \cite{kroemer2019review}. It remains surprisingly difficult to find robust and general solutions for this class of tasks without depending on specialised fixtures or other aids. Engineers have long devised ingenious methods for part insertion. However, these are either highly case-specific in manufacturing or not very robust in the face of hardware wear and tear or other environmental uncertainties. Humans solve such tasks without difficulties using visual or tactile perception, dexterous manipulation, and learning. This problem scenario is the focus of our work.
\begin{figure}
      \centering
      \includegraphics[width=0.5\textwidth]{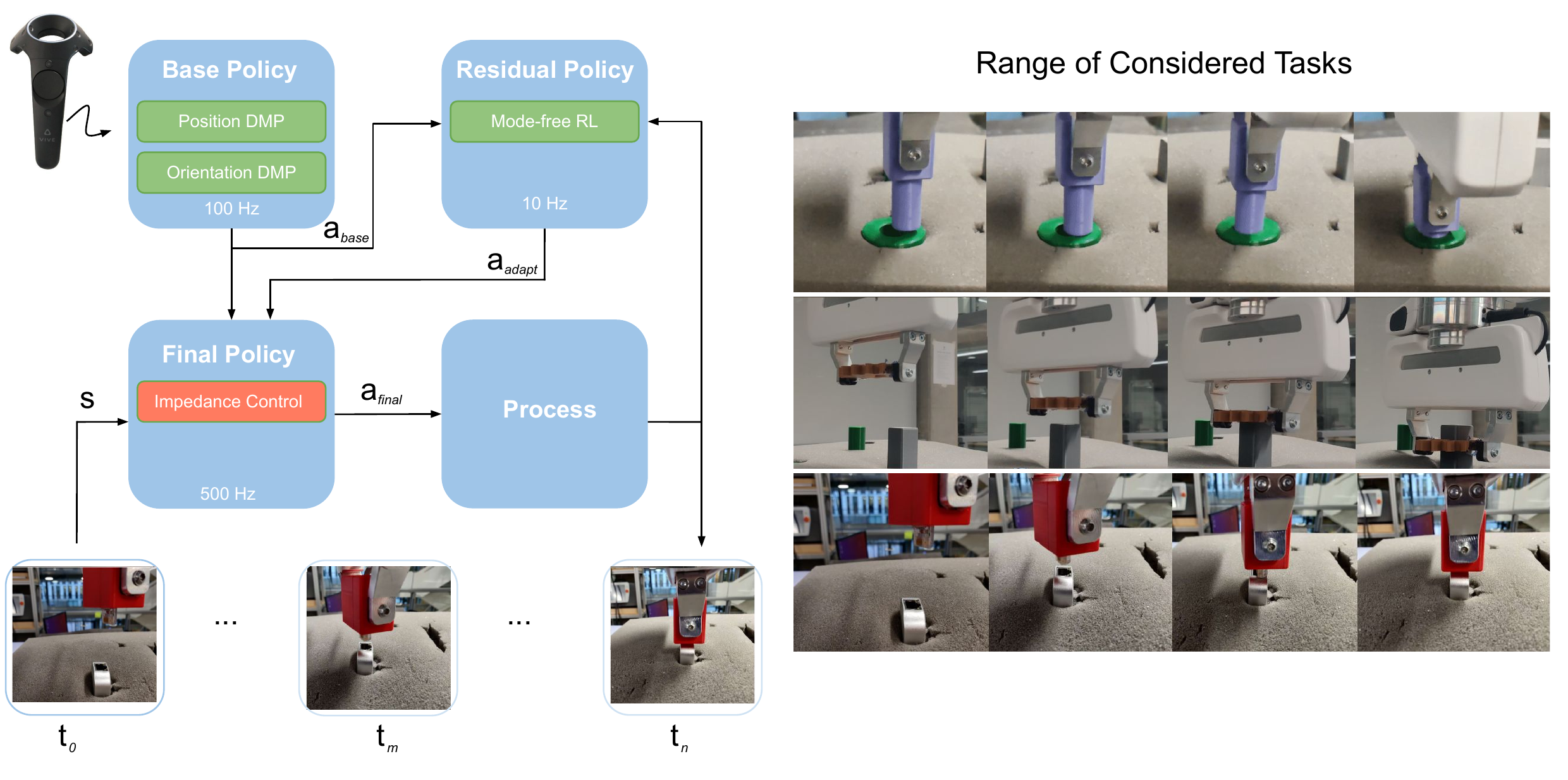}
      \caption{\small Left: outline of the proposed framework. 
      We collect demonstrations using an HTC Vive tracker,
      and extract an initial full pose policy using Dynamical Movement Primitives (DMPs, running at 100Hz). The control command produced by the DMP is corrected by an additional residual policy trained using model-free RL (run at 10Hz).
      The resulting motor command is then fed into a real time impedance controller (running at 500Hz) in a Franka Panda arm that performs peg, gear or LAN cable insertion in our physical setup. Right: The peg (top), gear (middle) and LAN cable (bottom) insertion tasks considered in this work.}
      \label{fig:architecture}
      \vspace{-5mm}
\end{figure}

Learning from demonstration (LfD)~\cite{schaal1997learning} is a popular approach for effective and rapid skill acquisition in physical robots. 
Existing solutions to LfD~\cite{sutanto2018learning, huang2020toward} have shown that robots can gain proficiency on tasks even when we may not have detailed models or simulations of complex environments. However, generalising to a range of scenarios without explicitly modelling and identifying the effects of contacts and friction remains difficult, hence inhibiting robot learning in these contact-rich settings. 
Model-free RL is beneficial for solving challenging tasks~\cite{vevcerik2017leveraging, luo2019reinforcement}, but requires large amounts of data and often impractical numbers of training episodes. Furthermore, many hours of training can result in higher levels of equipment wear and tear and increase the risk of more severe hardware damage due to the inherently exploratory nature of RL.

Roboticists have recognised the need for adaptive learning from demonstration schemes that occupy a middle ground~\cite{kroemer2019review}. This paper follows this spirit by combining LfD and model-free RL in a residual pose correction framework comprised of two complementing policies. DMPs~\cite{ijspeert2013dynamical} act as a base policy acquired by LfD. The DMP approach has always included an additive coupling term that allows for online modulation of a DMP policy -- i.e., DMPs are naturally interpreted as a form of residual learning. We instantiate this coupling term as a state-based policy using variants of state-of-the-art on-policy and off-policy RL trained in an online fashion (Figure~\ref{fig:architecture}, left). This yields a learning approach that is more sample-efficient and generalises to a range of start positions, orientations, geometries as well as levels of friction. We study the efficacy of this framework using peg, gear and LAN cable insertions - both in simulation and on a physical robot (Figure~\ref{fig:architecture}, right).

DMPs are ubiquitous in behaviour cloning settings and a popular technique for learning from demonstration. Although existing work has explored directly adapting DMP or coupling terms to improve task generalisation~\cite{peters2007applying, kober2009policy, peters2008reinforcement}, it is still unclear how best to do this for contact-rich insertion tasks. This paper explores DMP correction techniques and shows that a residual correction policy that adapts directly in task space, outside the canonical formulation, and uses reinforcement learning is the most effective strategy among all considered ways for learning insertion skills. Furthermore, this framework explicitly accounts for orientation-based policy corrections in task space. Our results show that orientation-based correction, which is typically not applied to existing residual learning frameworks such as~\cite{schoettler2019deep}, is essential for reliable and robust peg-in-hole insertion. This is partly challenging because additive orientations can introduce catastrophic singularities and locks, while formulating residual orientation-based corrections in quaternion space can be challenging to learn.

\textbf{Contributions}
The key contributions of this paper are:

\vspace{-4mm}
\Contributions{1}{An extensive comparison of the adaptation of different parts of the DMP formulation using a range of adaptation and exploration strategies for contact-rich insertion;}
\Contributions{2}{A framework for applying full pose residual learning on DMPs applied directly in task space, and the demonstration of its utility to three types of physical insertion tasks;}
\Contributions{3}{\rb{Showing that using full-pose residual nonlinear policies (e.g. RL-driven) to adapt DMPs results in more accurate, gentle and more generalisable DMP solutions.}}
\vspace{-4mm}
\section{RELATED WORK AND CONTRIBUTIONS}
\label{sec:related-work}
Insertion skills are essential in many robot manipulation applications. They are a crucial part of manufacturing assembly, home automation, laboratory testing and even surgical automation~\cite{kroemer2019review}. Existing literature on this problem can be organised into two broad categories based on whether or not the proposed methods split the task execution into phases based on contact times. We provide an overview of some of these methods and position our work in this context.

\textbf{Modelling Contacts}
It is common practice in extant approaches to first model the time(s) of contact and then to structure the insertion strategy in terms of two separate sub-problems - contact-state recognition and compliant control~\cite{kroemer2019review}. 
Compliant control generally relies on a human engineer to characterise the underlying friction and contact model~\cite{whitney1982quasi}. Recent approaches demonstrate high-precision assembly through the use of additional force information~\cite{inoue2017deep}. These methods require careful design of the manipulated objects and the external environment, e.g. assuming a flat surface surrounding the hole. These assumptions may not hold in many practical and day-to-day settings encountered in the real world, e.g. for RJ-45 connector insertion tasks (Figure \ref{fig:architecture}, bottom row). \ra{Also, it tends to be challenging to obtain accurate models of nonlinear changes in contact dynamics over time}, which affects reliability. In contrast, rLfD seeks to overcome these limitations by relying on strategies extracted from human demonstrations, and a learned policy that adapts and optimises the end-effector pose directly in task space.

\textbf{Learning from Demonstration (LfD)}
Learning policies for complex robot tasks can benefit from expert demonstrations \cite{havoutis2013motion}. LfD~\cite{schaal1997learning} is a widely adopted approach formalising this idea and has been applied previously in the context of peg-in-hole insertion tasks~\cite{zhu2018robot}, but does not account for model imperfections. DMPs \cite{ijspeert2013dynamical} have been shown to improve the generality of demonstrations in a variety of manipulation tasks \cite{sutanto2018learning, ude2014orientation, peters2007applying, kober2009policy, peters2008reinforcement, stulp2012reinforcement,gams2014coupling,likar2015adaptation,gams2014learning,abu2015adaptation,koutras2020dynamic}. However, these works do not consider full pose online adaptation behaviours in task space. 

In cases where the error signal is a linear combination of basis functions throughout movement executions, such as minimising force/torque feedback control error in bimanual manipulation tasks \cite{gams2014coupling, likar2015adaptation}, DMPs can be adapted by employing linear adaptive policies. Gams et al.~\cite{gams2014learning} directly modify the forcing term (i.e. adaption in parameter space) of a DMP iteratively and apply it to the task of wiping a surface. Abu-Dakka et al. \cite{abu2015adaptation} use fixed feedback models as constant gains and focus on adaptation by learning the error term with iterative learning control and adding it to the DMP trajectory. However, this does not target generalisation in the context of wide range of start configurations or complex geometries like RJ-45 connector insertion as we do in this work.

\begin{figure*}
      \centering
      \includegraphics[width=\textwidth]{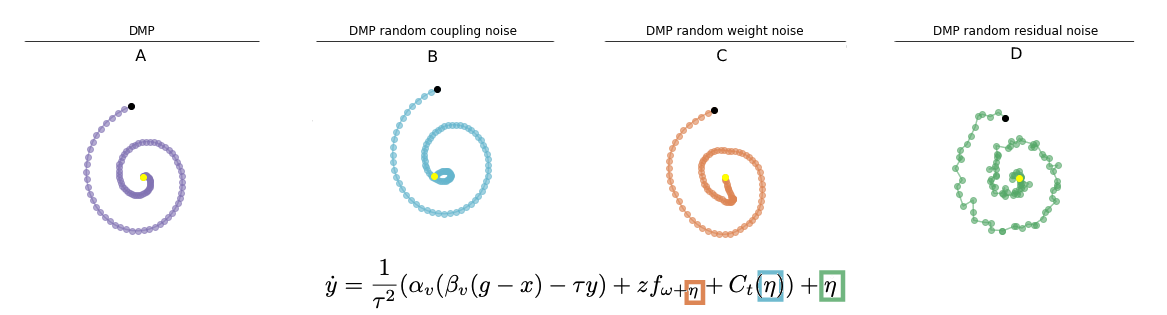}
      \caption{\small Types of exploration perturbations with Gaussian noise, $\eta$, for a simple Archimedian spiral. Applied on a translational DMP policy (purple - A), shown as equation above. Perturbing directly in task space (green - D) results in local exploration that is important for contact-rich manipulation. In contrast, perturbing the phase-modulated coupling term, $C_t$ (blue - B) or the parameters $\omega$ of the forcing term, $f_{\omega}$ (orange - C) results in locally smooth trajectories. Not perturbing the DMP is depicted in purple.
      }
      \label{fig:perturbations}
      \vspace{-5mm}
\end{figure*}

Previous work has also used Reinforcement Learning of feedback models, e.g. PoWER \cite{kober2009policy} or FDG~\cite{ peters2008reinforcement}, to learn or adjust nominal behaviours within a few iterations in parameter space of the DMP formulation. However, these approaches are limited to linearly weighted combinations of phase-modulated features, which are less expressive and can lead to unsatisfactory performance for contact-rich manipulation tasks. Alternatives, such as eNAC \cite{peters2007applying} have proposed adapting DMPs in the phase-based coupling term space instead, which is helpful for tasks that require reactive movement such as wiping~\cite{sutanto2018learning} or hitting a baseball \cite{peters2007applying}. This formulation does not allow for random local task space exploration (e.g. jiggling), which is helpful in contact-rich manipulation tasks (see Figure~\ref{fig:perturbations}). Instead, we propose to adapt DMPs directly in task space avoiding these limitations.

In cases where selecting arbitrary desired points of a trajectory can be done without affecting performance - such as in smooth surface painting or handover~\cite{huang2020toward}, both translational and orientation-based behavioural cloning (BC) policies can be adapted with an analytical approach. However, large perturbations of a robot's start configuration in the context of insertion often leads to dramatic differences between a demonstrated trajectory and the desired one. This can invalidate such modes of teaching due to the significant distance between both trajectories. Instead, we propose a generalisation of such methods' structure and describe it as a framework that allows learning to adapt DMPs directly in task space. We show that this allows us to rely on nonlinear model-free approaches, which can benefit contact-rich insertions. Our results show that using this residual learning formulation leads to a robust to perturbations and sample efficient solution.

\textbf{Residual RL}
Acknowledging the difficulties of running RL on physical systems, approaches like~Johannink et al.~\cite{johannink2019residual} combine conventional contact model-based control with model-free reinforcement learning but do not handle orientation-based corrections. Instead, they focus on sliding at low levels of friction. Schoettler et al.~\cite{schoettler2019deep} propose using stronger priors and combine SAC with a proportional controller to solve industrial position-only insertion tasks. They do not scale to full pose corrections and encourage the use of higher forces via a dense reward function. This works well in Cartesian space settings with very low perturbations but is limiting in the context of insertion. Tuning such dense rewards requires additional insights and potentially increases the risk of hardware damage. 
\rb{Zeng et al. \cite{zeng2020tossingbot}, employ adaptive policies for learning how to throw. However, none of these works shows how best to apply residual learning to DMPs, a broadly applicable and flexible class of models of significant interest in robotics. Our extensive evaluation sets out to answer this.}

\textbf{Summary}
It is clear that there has been substantial work on DMP adaptation using residual policies. However, it remains unclear how best to do so for contact-rich insertion tasks. This paper explores this and finds that a framework bridging reinforcement learning and DMP learning from demonstration in task space (rLfD) is most effective when paired with orientation aware corrections. This framework is evaluated in both simulated and physical systems, running under real-time constraints. Results show that the proposed formulation significantly reduces the sample requirements during training and allows for the use of a sparse reward while preserving the overall improved accuracy achieved by model-free RL. In practice, this is important as it limits the risk of hardware damage and thus makes this approach feasible for real-world applications. 
\vspace{-3mm}
\section{METHODS}
\label{sec:methods}
In this paper, we learn a base policy, $\pi_{b}$, from expert demonstrations, represented directly in task space. We extract separate policies for each individual task in (Figure~\ref{fig:architecture}, right). We use two separate formulations for $\pi_{b}$ - namely translational and orientational DMPs~\cite{ijspeert2013dynamical, ude2014orientation}. The learned behaviour is then executed by a suitable robot controller, i.e. in position or impedance control, to produce the final robot motion. The translation-based base policy $\hat{\pi}_b$ is described by position and velocity terms, $(\textbf{x}, \dot{\textbf{x}})$, in 3D Cartesian space. The orientation-based base policy, $\tilde{\pi}_b$ is described using the orientation in quaternion and angular velocity, $(\textbf{q}, \boldsymbol{\omega})$. The resulting base policy, $\pi_{b}$ is capable of imitating the full pose trajectory of the demonstrated behaviour. Such a formulation can generalise well to perturbations of the initialisation and timing conditions, especially when executed in free space. However, it cannot adapt well to previously unseen task settings and environmental perturbations, which are typical for contact-rich manipulation tasks. In this work, we study how to alleviate this limitation using an additional residual policy, $\pi_{\theta}$ by employing state-of-the-art model-free RL that can act directly on the end-effector twist.
\vspace{-2mm}
\subsection{Problem Formulation}
We consider a finite-horizon discounted Markov decision process (MDP), $\mathcal{M} = (S, A, P, r, \gamma, T)$ with transition probability $P:S\times A \times S \mapsto [0, 1]$ and a reward function $r(x, a)\in\mathbb{R}$. Let $x \in S \subseteq \mathbb{R}^{n_x}$ be an element in the state space $S$, and $a \in A \subseteq \mathbb{R}^{n_{a}}$ denote the desired robot end-effector velocity (action). 
Then, a stochastic control policy parameterised by $\theta$, $\pi_{\theta}(a|x)$ defines the probability of selecting an action $a$ given a state $x$. Let $ \zeta = \left(\textbf{s}_{t_0}, \ldots, \textbf{s}_{T}\right)$ be a trajectory with a discrete horizon $T$, a discount function $\gamma(\cdot)$ and a state-action tuple $s_t = (x_t, a_t)$ and a trajectory return $R(\zeta) = \sum_{t=t_0}^{T}\gamma r(x_{t}, a_{t})$. In this context, we can define an optimal policy as 
  $\pi^{*}=\arg\max_{\pi \in \bar{\pi}}J(\pi)$,
where $J(\pi) = \mathbb{E}_{s_0 \sim p(s_0), a_t\sim \pi(s)} [R(\zeta)]$ and $\bar{\pi}$, the set of all policies.

We can use a policy gradient method to optimise the objective from the sampled trajectories. By relying on a DMP to extract a fixed and pre-computed offline continuous base policy, $\pi_{b}$, we significantly reduce the complexity of the task for $\pi_{\theta}$.
The residual policy has to learn how to deviate from the base policy by correcting model inaccuracies and potential environmental changes during execution. 
The final policy can compensate for system uncertainties through an adaptive term obtained from $\pi_{\theta}$ while using a base policy, $\pi_{b}$ acquired through human demonstrations. A core question in this context is what part of the DMP formulation should $\pi_{\theta}$ be applied to. We propose to adapt base policies directly in task space as described in the following subsections.
\vspace{-2mm}
\subsection{Translation-based Residual Corrections for DMPs}
Contact-rich manipulation has inherent non-linear dynamical effects between the robot and the surrounding environment. 
In most insertion tasks, slight perturbations can have drastic consequences on the performance of $\pi_{b}$. 
To this end, utilising model-free solutions can help relax the challenge of modelling contact dynamics.

Adapting DMP formulations is a well known approach as discussed in Section~\ref{sec:related-work}. Consider a point-to-point movement with a DMP defined as, $\dot{y} = \frac{1}{\tau^2}(\alpha_v (\beta_v (g - x) - \tau y) + f_{\omega} + C_t)$. In this context, an online adaptation term can be learnt by using exploration noise $\eta$ that is applied to different components of this equation, e.g. to the parameters, $\omega$ of the forcing term $f_{\omega}$, the phase-modulated coupling term $C_t$ or outside the phase modulated DMP formulation (e.g. in task space). For brevity, consider describing these different ways of applying $\eta$ with colour-coding, and namely
\begin{equation}
  \dot{y} = \frac{1}{\tau^2}(\alpha_v (\beta_v (g - x) - \tau y) + zf_{\omega + \forcing{\eta}} + C_t(\coupling{\eta})) + \task{\eta}.
    \label{eqn:cart_dmp}
\end{equation}
Most existing work focuses on adapting nominal behaviours by learning a forcing term with some exploration signal $\eta$, applied in parameter space (shown in orange) \cite{peters2008reinforcement, kober2009policy, abu2015adaptation, gams2014learning}. At the same time, some have studied the effects of learning a phase-modulated coupling term\footnote{We will refer to this as adapting in coupling term space to differentiate between additive terms.} (shown in blue) \cite{peters2007applying, sutanto2018learning}. Recent work, \cite{huang2020toward} uses adaptive analytical models that can be applied directly in task space, too (shown in green). All of these different modes of exploration lead to learning different residual terms that have their benefits. 
However, very little has been done to compare the different residual signals that are learnt using the different colour-coded exploration noise from Equation~\ref{eqn:cart_dmp}. 
Similar in spirit to \cite{Sehnke:2010gi}, this work studies the benefits of those different adaptive formulations and proposes correcting directly in task space. We conjecture that exploration in task space (Figure~\ref{fig:perturbations}.D) encourages jiggling motion strategies that are useful in the context of insertion. A learnt policy that adapts in task space should thus be superior for contact-rich insertions.
\paragraph{Residual Learning in Task Space}
In light of our conjecture, we propose to adopt recent advances of residual learning~\cite{johannink2019residual,schoettler2019deep} and extract a residual translational policy $\hat{\pi}_{\theta}$, parameterised by $\theta$ that adapts a base prior policy, $\hat{\pi}_{b}$ outside of the canonical system and directly in task space. The final translational policy can then be defined as $\hat{\pi}_{f}=\hat{\pi}_{b}+\hat{\pi}_{\theta}$. A key aspect of this formulation is that $\hat{\pi}_{\theta}$ is a complementary policy that operates in task space alongside the DMP policy $\pi_{b}$. We now extend this formulation to orientation-based DMPs.
\vspace{-2mm}
\subsection{Generalising to Full Pose Corrections}
\label{sec:pose_corrections}
Given environmental uncertainties and assuming also some inaccuracies in the robot control itself, the translational residual formulation introduced above can correct this by combining an adapting policy $\hat{\pi}_{\theta}$ using the DMP as a prior policy $\hat{\pi}_{b}$, resulting in 
$\hat{\pi}_{f} = \hat{\pi}_{b} + \hat{\pi}_{\theta}$.
\paragraph{Orientation-based Corrections}
The additive correction introduced above is unsuitable for orientations due to the risk for singularities or locks associated with Euler representations. Assuming fixed orientations or constant angular velocity could, in principle, relax this constraint. However, such approximations restrict applicability as motion is rarely fixed in orientation by nature. In order to address this, we introduce a residual formulation capable of accommodating orientations. Quaternion based representations lead to smooth interpolations that are compact and do not suffer from Gimbal locks. Therefore, we propose the following orientation-based residual framework.
\paragraph{Orientation based residual corrections in Task Space}
In this context, a normalised quaternion $Q = [q^w, q^x, q^y, q^z ]$ such that $||Q|| = 1$ is defined as a vector with a real scalar value $q^w$ and a vector $[q^x, q^y, q^z]$ of imaginary components. \ra{}\rb{We define composition $\circ$ between two quaternions using Shuster's notation~\cite{shuster2008nature}. 
An orientation $Q_f$ along some orientation trajectory is produced from a policy $\tilde{\pi}$. In this work, $\tilde{\pi}$ is composed of two separate policies: a stochastic residual orientation policy $\tilde{\pi}_{\theta}$ which operates in task space and a base orientation policy $\tilde{\pi}_b$ from an orientational DMP~\cite{ude2014orientation}}.
\paragraph{Learning residual corrections in quaternion space}
Residual orientation is composed in quaternion space and adapts the orientation of the end-effector. The policy predicts the parameters of an angle-axis representation which consists of a 3D unit vector $\textbf{r}$ around which the robot end-effector is rotated by a scalar angle $\alpha$. This results in a residual orientation vector $\{\alpha, \textbf{r}\}$. Assuming that $\alpha \in [-\pi, \pi]$, which covers all rotations, it follows that $cos(\alpha/2)$ is strictly positive. Then, a correction $Q_{\Delta} = \{q_{\Delta}^{w}, q_{\Delta}^{x},q_{\Delta}^{y},q_{\Delta}^{z}\}$ can be computed where $q_{\Delta}^{w} = cos(\alpha/2)$ is the real part of the quaternion. The orientation-based adaptation relative to the base orientation can be described as quaternion, \ra{\cite{grubin1970derivation}} by
\begin{equation}
    Q_{\Delta} = [cos(\alpha/2),\ \frac{\textbf{r}}{||\boldsymbol{r}||} sin(\alpha/2)], 
\end{equation}
where $||\boldsymbol{r}||$ is the L2 norm of the rotation vector. This allows us to obtain a quaternion correction term using real numbers. Then, adding the correction to the prior orientation is achieved with quaternion multiplication as 
\begin{equation}
Q_{f} = Q_{\Delta}\circ Q_{b}.
\end{equation}
The final quaternion $Q_{f}$ is then converted to angular velocity using a log transform \cite{ude2014orientation}.
\paragraph{A Framework for Full Pose Residual Corrections}
In summary, we propose a complete, twist policy $\pi_{f}$ as
\begin{equation}
\pi_{f} = [
           \hat{\pi}_{f},
        \tilde{\pi}_{f}
         ]^\text{T}.
\end{equation}
In practice, we combine the work introduced in \cite{huang2020toward} and \cite{schoettler2019deep} and extend it to a general framework for full pose residual learning. The individual components of our policy can be extracted independently from each other and can be of different nature. They can be learnt (e.g. by using RL) or analytically defined (e.g. by using recursive least squares (RLS) or similar to \cite{huang2020toward}). Like ~\cite{schoettler2019deep} we complement a base policy with a residual component. However, instead of using a basic proportional controller, we focus on improving DMPs for contact-rich tasks. Residual learning in task space happened to work best. We extend this by adding residual corrections and control with RL in real-time.
\vspace{-3mm}\subsection{Execution Details}\vspace{-1mm}
\paragraph{DMPs} We used a single demonstration to build the base policy. This was sufficient to extract 100\% successful insertions when an episode starts up to 3mm away from that starting position. We used 40 and 70 basis functions for $\hat{\pi}_{b}$ and $\tilde{\pi}_b$ respectively. These values were chosen from a grid search of parameter sizes (see web page for details).
\paragraph{Model-free Policies}

\textbf{SAC}:
\label{appdx:sac}
We found that this off-policy approach requires a recurrent neural policy to work in the lower frequency regime given the higher frequency controller. We used a recurrent policy with a cell state of 40. The actor policy comprised 400 and 300-dimensional feed-forward layers with ReLU activation functions. The critic had a single feed-forward layer of size 300. Those parameters were chosen in a grid search across policy network sizes. We used 32 policy update steps per iteration.

\textbf{PPO}
We utilise the clipped objective as in Schulman et al.~\cite{schulman2017proximal}. We used a curriculum for the physical experiments to reduce the sample requirements by varying the starting configuration. \ra{As the agent improves we increase task complexity until it becomes 1.5 cm away from the start configuration of the demonstration.} PPO was less sensitive to network sizes.
\paragraph{Implementation Details}
Both environments are built using Tf-Agents~\cite{TFAgents}. Training is performed using a sparse reward. All policies rely on a normal projection for continuous actions, which uses $\tanh$ to limit its output.

We use MuJoCo~\cite{todorov2012mujoco} and SL~\cite{schaal2009sl} for experimenting in simulation and the physical world, respectively. In simulation, we utilise Robosuite~\cite{robosuite2020} and use their position controller applied on low-level actions obtained with inverse kinematics. To evaluate rLfD in the physical world, we used a Jacobian transpose Cartesian impedance control run in real-time (details on the website). Both controllers run at 500Hz, while $\pi_{b}$ at 100Hz and $\pi_{\theta}$ at 10Hz (see Figure~\ref{fig:architecture}).

The framework supplies set-points to a real-time running controller and uses a real-time clock to synchronise the concurrently running modules. We extrapolate the DMP corrections using a standard fifth-order polynomial while assuming static repetition of the produced residual term.
\vspace{-2mm}
\section{EXPERIMENTAL EVALUATION}\vspace{-1mm}
\label{sec:experimentation}
We evaluate the effects of using different adaptive corrections in the context of skill acquisition for manipulation. We perform a sequence of empirical evaluations using a 7 DoF Franka Emika Panda arm. 
Our results show that rLfD with model-free reinforcement learning techniques is both sample efficient and improves the performance of DMPs on contact-rich manipulation tasks. We split this section into two main parts: studying the effects of using residual corrections on a simulated environment and applying rLfD in the physical world on three different complexity insertion tasks.
\vspace{-3mm}
\subsection{Applying Residual Corrections}
\label{sec:applying_corrections}
Next, we conduct a thorough study in simulation. We apply different types of corrections to a DMP in the context of contact-rich manipulation.
\paragraph{Experimental Details}
\begin{figure}
    \vspace{2mm}
      \centering
      \includegraphics[width=0.45\textwidth]{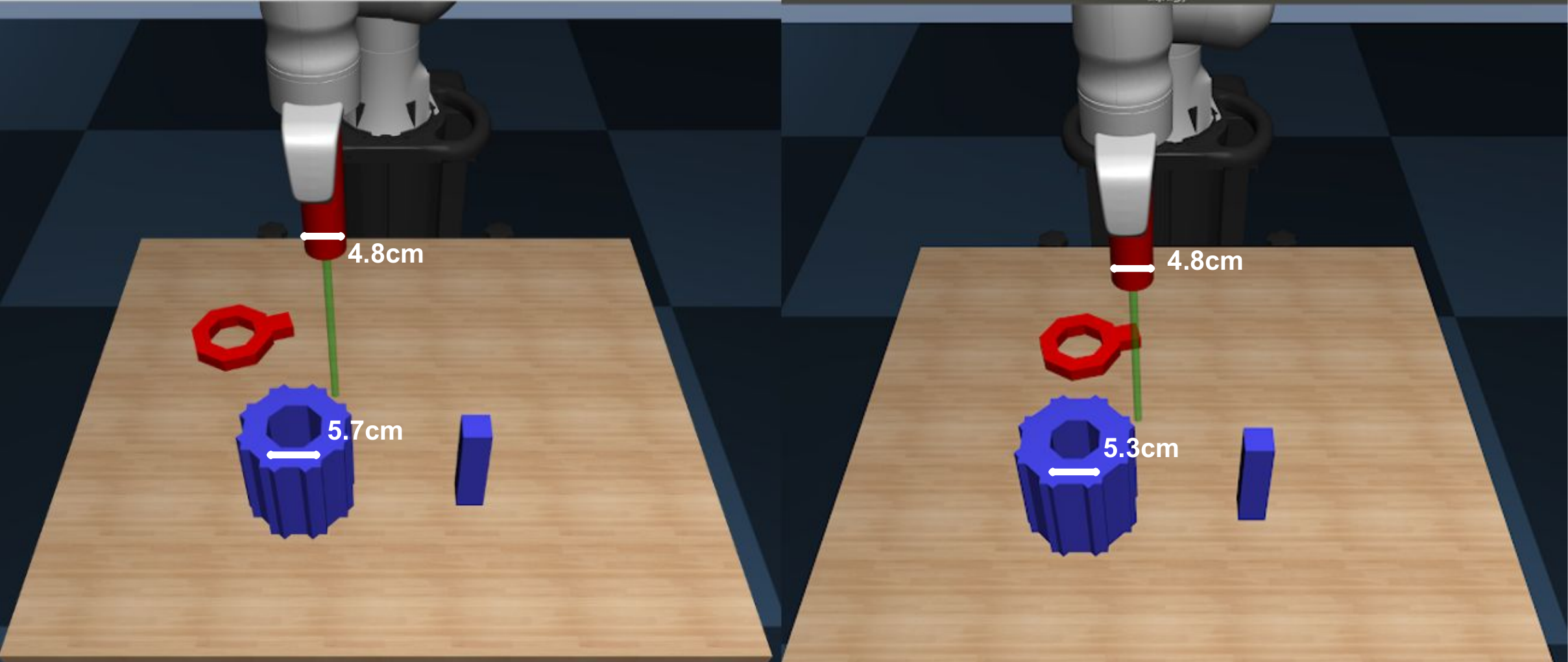}
      \caption{\small An easy task (left) and a harder task (right). The robot is initialised with a position sampled uniformly within $\pm$12cm along all axes of the initial position of the demo. Difficulty is defined by the size of the hole. A task is complete when a peg is fully inserted.}
      \label{fig:simulation}
    \vspace{-5mm}
\end{figure}
\begin{figure*}
      \centering
      \includegraphics[width=\textwidth]{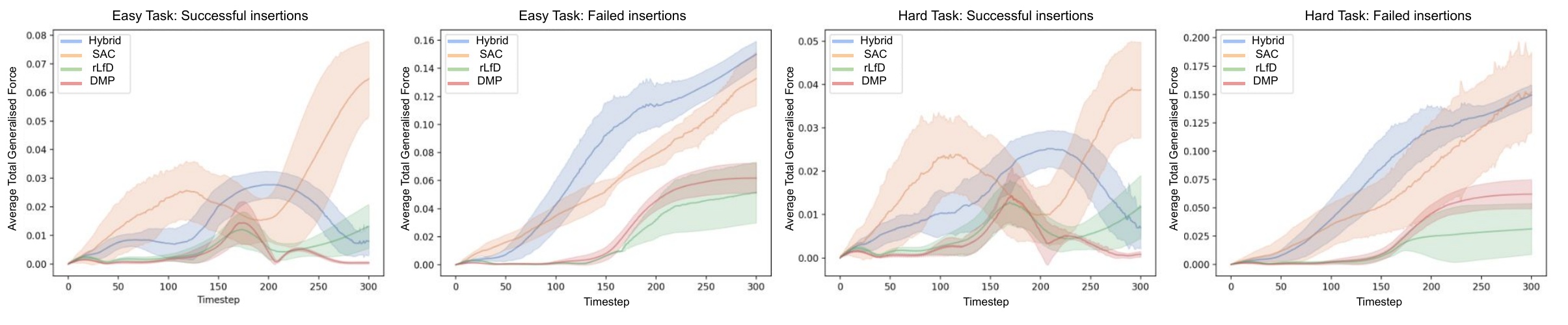}
      \caption{\small Comparison between using residual, hybrid, DMP and model-free policies. The residual policy (green) consistently results in experiencing generalised forces comparable to the forces experienced using only a DMP (red) when succeeding. It experiences even smaller forces across both the easy and hard tasks when failing to insert. Lower is better.}
      \label{fig:res_vs_mfree}
      \vspace{-5mm}
\end{figure*}
We study the utility of applying residual corrections in the context of learning to insert in simulation. We utilise two separate tasks we refer to as 'easy' and 'hard'. The difficulty of a task is defined by the size of the hole a peg has to be inserted in (see Figure~\ref{fig:simulation}). A tighter hole indicates more difficulties, such as potential jamming due to friction during insertion. The task with a larger hole could be solved using only translation-based corrections. Here, we train on the easy task but evaluate on both.
\begin{table}[!htpb]
\Huge
\caption{\small Accuracy of applying exploration during learning on different parts of the DMP formulation (Type colour from Fig.~\ref{fig:perturbations}). Eff. is efficiency- the number of episodes a model was trained for.}
\vspace{-5mm}
\begin{center}
\resizebox{0.49\textwidth}{!}{
\begin{tabular}{|c|c|c|c|c|c|c|c|}
\hline
\rbno{Type} & Exploration Type & \rano{Model} & Easy & Hard & Average & Eff. & Reward \\
\hline
\hline
A & \dmp{No corrections} & DMP \cite{ijspeert2013dynamical} & 24.0\% \huge{$\pm{2.5}$} & 8.0\% \huge{$\pm{1.4}$} & 16.0\% \huge{$\pm{2.0}$} & n/a & n/a\\
\hline
B & \coupling{phase modulated coupling} & eNAC~\cite{peters2007applying} & 7.2\% \huge{$\pm{1.9}$} & 3.4\% \huge{$\pm{2.2}$} & 5.3\% \huge{$\pm{2.1}$} & 8K & $exp\{-L_1\}$ \\
\hline
 C & \forcing{forcing-term parameters} & FDG~\cite{peters2008reinforcement} & 23.6\% \huge{$\pm{4.3}$} & 16.2\% \huge{$\pm{1.9}$} & 19.9\% \huge{$\pm{3.1}$} & 8K & $exp\{-L_1\}$\\
\hline
 C & \forcing{forcing-term parameters} & PoWER~\cite{kober2009policy} & 32.2\% \huge{$\pm{2.8}$} & 14.4\% \huge{$\pm{2.7}$} & 23.3\% \huge{$\pm{2.8}$} & 8K & $exp\{-L_1\}$\\
\hline
D & \textbf{\task{task space translation}} & \textbf{rLfD (ours)} & \textbf{94.8\% \huge{$\pm{1.3}$}}  & \textbf{55.0\% \huge{$\pm{2.7}$}} & \textbf{74.9\% \huge{$\pm{2.0}$}} & \textbf{700} & $\mathbb{1}[L_2 \leq \kappa]$\\
\hline
\end{tabular}}
\end{center}
\label{tabl:correcting}
\vspace{-4mm}
\end{table}
\paragraph{Adapting different parts of a DMP}
We summarise the effect of using a range of RL approaches that apply exploration noise during learning to different components of the movement primitives formulation \ra{(see website for baseline details)}. Our findings, reported in Table~\ref{tabl:correcting}, further confirmed the conjecture from Figure~\ref{fig:perturbations} that exploring directly in task space is more effective for learning in terms of accuracy. As a result, the learnt task corrections can achieve more sample efficient solutions using only sparse rewards, which is of essential importance to applying RL in practical settings.
\paragraph{Adaptive strategy selection in task space}
\label{sec:adaptive_strategy}
Choosing what part of a DMP to explore to train an RL agent is essential to their successful application in contact-rich settings. However, it is unclear whether using a nonlinear RL-based adaptive strategy is, in fact, necessary for contact-rich insertions. We compare using two different types of nonlinear correcting strategies to a recursive least-squares linear policy \cite{kumar2016optimal} and random adaptive noise \cite{pottle1963digital} applied directly in task space. Results (Table~\ref{tabl:correcting_strategies}) show that nonlinear adaptive strategies are, in fact, helpful in the context of adapting DMPs for contact-rich manipulation tasks.
\begin{table}[!htpb]
\Huge
\caption{\small Residual adaptive policies in task space for DMPs (Exploration type colour from Fig.~\ref{fig:perturbations}). Insertion accuracy table.}
\vspace{-2mm}
\begin{center}
\resizebox{0.48\textwidth}{!}{
\begin{tabular}{|c|c|c|c|c|c|c|c|}
\hline
\rbno{Type} & Corrections & \rano{Adaptive Policy} & Easy & Hard & Average & Eff. & Reward \\
\hline
\hline
A & \dmp{translation} & Random \cite{pottle1963digital} & 25.8\% \huge{$\pm{4.3}$} & 9.0\% \huge{$\pm{1.8}$} & 17.4\% \huge{$\pm{3.1}$} & n/a & n/a\\
\hline
A & \dmp{translation} & Linear~\cite{kumar2016optimal} & 25.4\% \huge{$\pm{3.6}$} & 8.0\% \huge{$\pm{2.6}$} & 16.7\% \huge{$\pm{3.1}$} & n/a & n/a\\
\hline
D & \textbf{\task{translation}} & \textbf{SAC} & \textbf{94.8\% \huge{$\pm{1.3}$}}  & 55.0\% \huge{$\pm{2.7}$} & 74.9\% \huge{$\pm{2.0}$} & \textbf{700} & $\mathbb{1}[L_2 \leq \kappa]$\\
\hline
D & \textbf{\task{translation}} & \textbf{PPO} & 87.6\% \huge{$\pm{1.5}$} & \textbf{69.0\% \huge{$\pm{4.6}$}} & \textbf{78.3\% \huge{$\pm{3.1}$}} & 25.5K & $\mathbb{1}[L_2 \leq \kappa]$\\
\hline
\end{tabular}}
\end{center}
\label{tabl:correcting_strategies}
\vspace{-4mm}
\end{table}
In all cases, only the DMP policy is used for the first half of an episode, with both the base and residual policies for the rest of the execution. While using PPO~\cite{schulman2017proximal} performs better both on average and in terms of generalising to more challenging tasks, it requires a higher number of steps to reach this level of performance. 
In contrast, SAC~\cite{haarnoja2018soft} can produce similar results in only a fraction of the required steps \ra{but requires an increased number of update iterations and thus a longer time to train}. The baselines perform worse on average but require less training as they did not rely on iterative exploration.
\paragraph{Utilising nonlinear policies applied in task space}
Nonlinear policies can significantly improve DMPs performance for contact-rich insertion tasks as shown above. In this section we compare rLfD against a hybrid and an RL-only solution applied in task space. We report results in Table~\ref{tabl:styles_of_adaptation}.
\begin{table}[!htpb]
\Huge
\caption{\small Comparing different ways of using nonlinear policies (Exploration type colour from Fig.~\ref{fig:perturbations}). Insertion accuracy table.}
\vspace{-2mm}
\begin{center}
\resizebox{0.48\textwidth}{!}{
\begin{tabular}{|c|c|c||c||c|c|c|c|}
\hline
\rbno{Type} & Corrections & \rano{Policy Type} & Easy & Hard & Average & Eff. & Reward \\
\hline
\hline
A & \dmp{None} & DMP & 24.0\% \huge{$\pm{2.5}$} & 8.0\% \huge{$\pm{1.4}$} & 16.0\% \huge{$\pm{2.0}$} & n/a & n/a\\
\hline
D & \task{translation} & (\rbno{pure}) SAC~\cite{haarnoja2018soft} & 94.4\% \huge{$\pm{1.2}$} & 41.8\% \huge{$\pm{7.2}$} & 68.1\% \huge{$\pm{4.2}$} & 12.5K & $-(\alpha * L_{1} + \frac{\beta}{L_{2}-\epsilon})$\\
\hline
D & \task{translation} & (hybrid) SAC & 57.2\% \huge{$\pm{2.5}$} & 45.6\% \huge{$\pm{2.7}$} & 46.4\% \huge{$\pm{2.6}$} & 8K & $\mathbb{1}[L_2 \leq \kappa]$ \\
\hline
D & \textbf{\task{translation}} & \textbf{(rLfD) SAC (ours)} & \textbf{94.8\% \huge{$\pm{1.3}$}}  & \textbf{55.0\% \huge{$\pm{2.7}$}} & \textbf{74.9\% \huge{$\pm{2.0}$}} & \textbf{700} & $\mathbb{1}[L_2 \leq \kappa]$\\
\hline
\end{tabular}}
\end{center}
\label{tabl:styles_of_adaptation}
\vspace{-4mm}
\end{table}

\begin{figure*}
    \begin{subfigure}{0.36\textwidth}
        \includegraphics[width=1.0\textwidth]{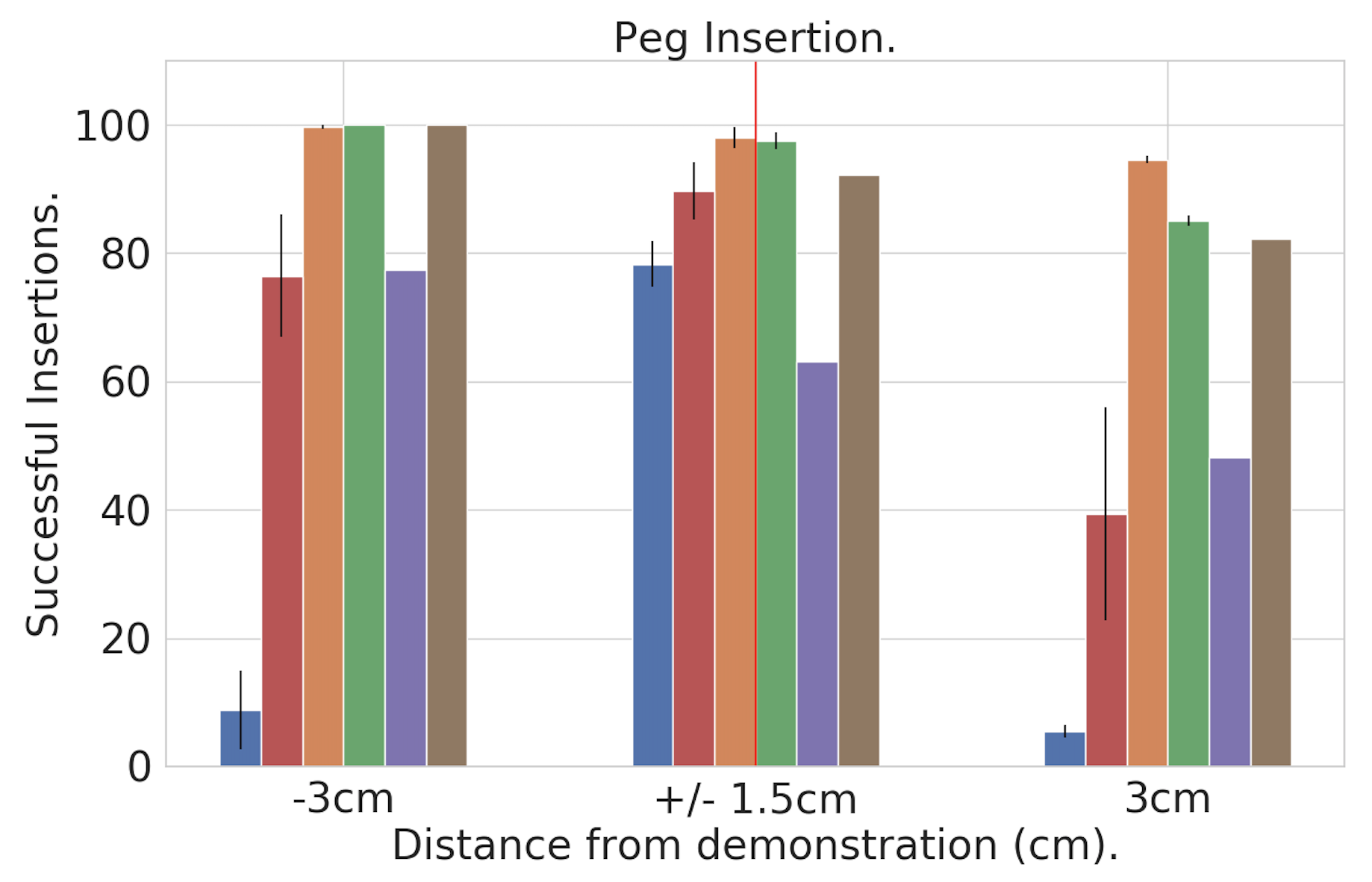}
    \end{subfigure}
    \begin{subfigure}{0.31\textwidth}
        \includegraphics[width=1.0\textwidth]{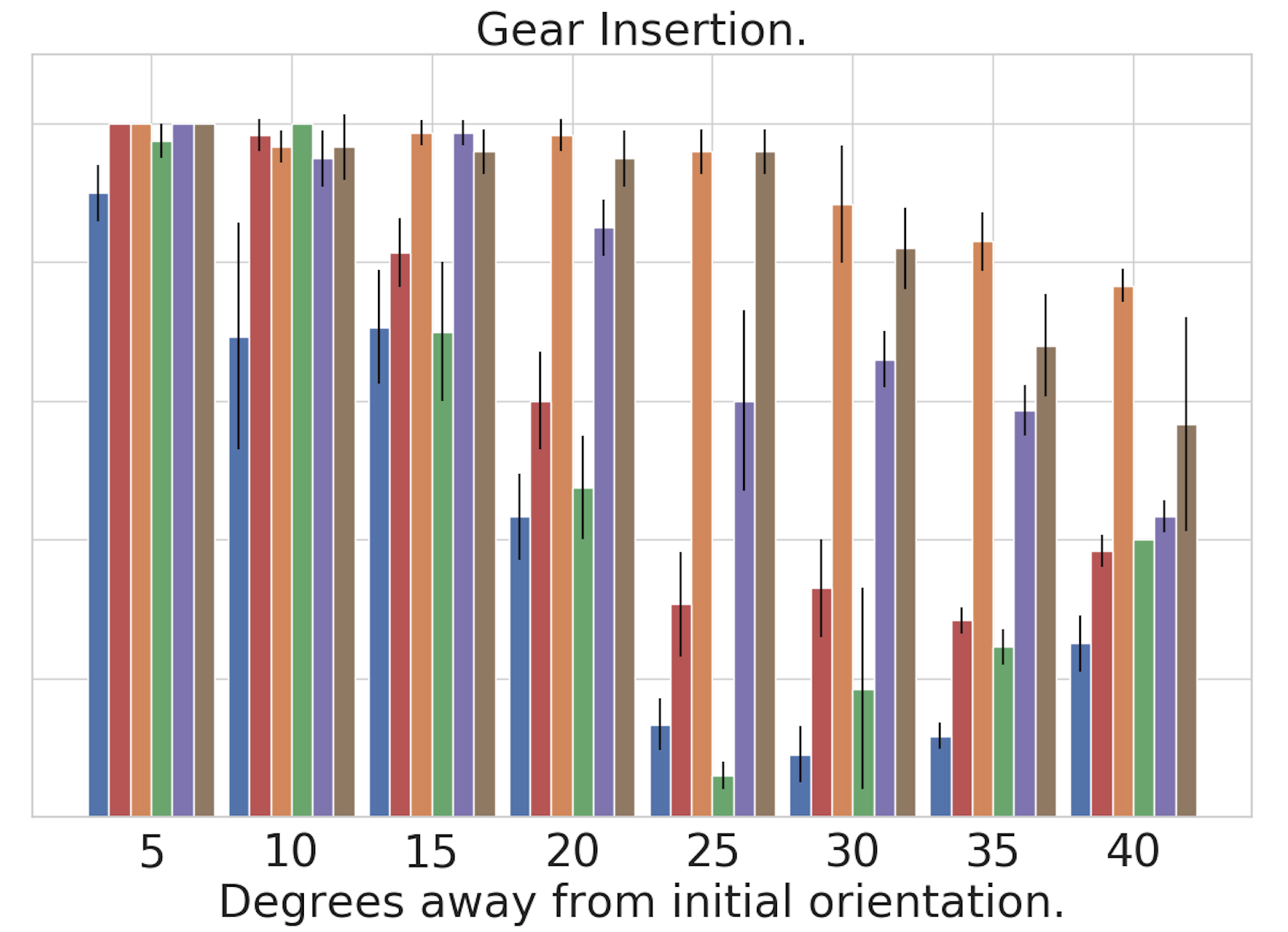}
    \end{subfigure}
    \begin{subfigure}{0.31\textwidth}
        \includegraphics[width=1.0\textwidth]{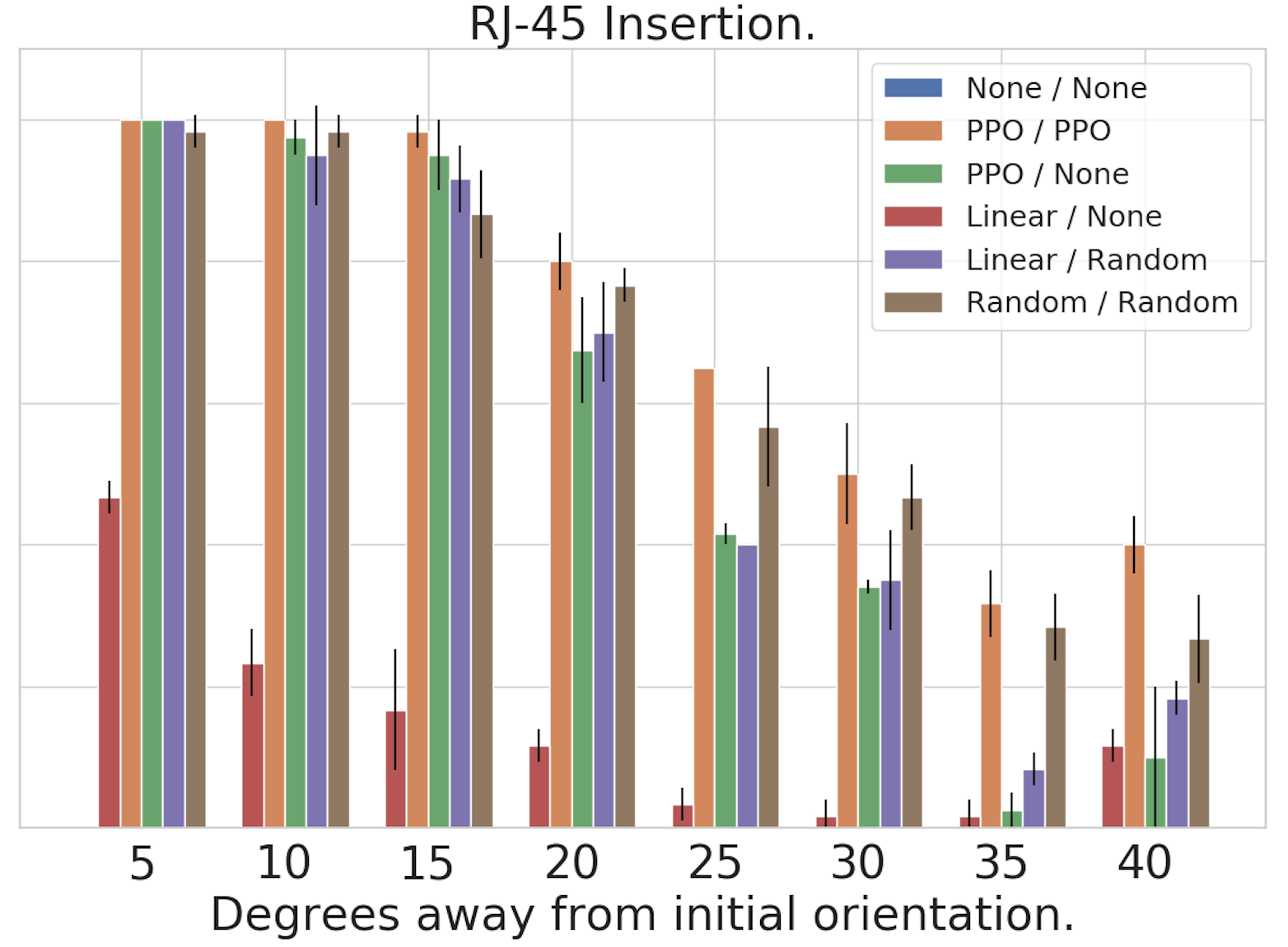}
    \end{subfigure}
    \caption{\small \rano{Successful peg, gear and RJ-45 insertions. Peg insertion results along the x axis are cm away from the initial position (illustrated by a red vertical line). Gear and RJ-45 results are plotted along the x axis as degrees away from the initial orientation. Higher is better.}}
    \vspace{-0.4cm}
\label{fig:pos}
\end{figure*}
Using a model free RL solution is known to be appealing due to its tremendous success in improving accuracy \cite{haarnoja2018soft}. Our findings confirm this too. \rb{Table~\ref{tabl:styles_of_adaptation} shows that using just SAC with no prior $\pi_{b}$ can get comparable performance to rLfD. However, a pure RL agent requires carefully engineered rewards and significantly longer training to reach that level of performance.} We used $\alpha=10$, $\beta=0.002$, $\epsilon=0.0001$ for the engineered dense reward case reported in row 2 of Table~\ref{tabl:styles_of_adaptation}. \rb{In contrast, rLfD generalised better to the harder task and allowed for a sparse reward function using a fraction of the training requirements.}

Using a hybrid formulation is a promising related approach. Lee et al.~\cite{lee2020guided} learn policies from visual pre-trained state-space representations with SAC by switching between model-based and model-free policies. We compare against a similar switching strategy to the proprioceptive state representations used by rLfD, switching between a base policy (learned via DMP) and a model-free policy. Using hybrid switching is appealing as it does not require prior knowledge of combining RL with a DMP formulation. In our tasks, we found switching to perform worse than rLfD. This indicates that rLfD may perform well if the more informative, visual representations proposed in Lee et al.~\cite{lee2020guided} are included. We leave this study for future work.

\paragraph{Gentle part insertion, analysis}
Our analysis (Figure~\ref{fig:res_vs_mfree}) shows that relying on SAC in both hybrid (blue) and purely model-free (orange) approaches results in a more forceful solution when compared to rLfD (green) \ra{possibly due to the larger magnitude of the action space required to compensate for the lack of a base policy}. The DMP (red) failed at times by constantly pushing downwards, which contrasts with the less forceful searching of rLfD (green).
\vspace{-2mm}
\subsection{Pose Corrections with Concurrent Real-time Control}
\label{sec:hardware}
In this subsection, we scale rLfD to correct the full pose in a sequence of physical insertion tasks using a less precise impedance controller ran in real-time.
\paragraph{Tasks Description}
We consider three tasks of differing complexity - peg insertion, similar in setup to the simulated tasks above; gear; and RJ-45 connector insertions (see Figure~\ref{fig:architecture}). Each of those tasks introduces an additional mode of complexity within the context of insertion. While physical peg insertion poses challenges due to the friction-heavy interactions with the highly nonlinear surrounding world, it can be solved by mainly relying on translational corrections. On the other hand, the gear insertion task requires perfect alignment of the squared peg with the hole. Due to the tight fit, it also requires a certain amount of downward force to achieve insertion. Finally, inserting an RJ-45 connector requires both perfect position and orientation. Moreover, the connector's plastic tip requires that a precise amount of force be applied to avoid breaking, further increasing task complexity (webpage for more details).

\paragraph{Experimental Details}
\label{sec:general_rpc}

In this subsection we provide additional experimental setup details for the tasks we use. 

\textbf{Peg Insertion} The peg's width, depth and height are $28\times28\times77$mm with a hole with 0.4mm clearance. Evaluation was done on 500 starting points that were sampled randomly from six, 3D uniform distributions, each centred at up to +/- 3cm away along each axis from the demonstrated position.

\textbf{Gear and Lan Insertion}
The gear is of size $79.2\times79.85\times10.79$mm with a square hole of $23\times23\times10$mm. Lan cable is standard RJ-45. Evaluation uses 160 uniformly sampled unit vectors per task, split into 8 bins of 20 samples of up to 40$^{\circ}$ and $\pm 1cm$ away from the demo orientation.

\textbf{Execution Details}
Each episode lasted at most 10 seconds. We used $\pi_{\theta}$ after 3.9 seconds of executing the base policy.

\paragraph{Using full pose corrections}
\label{sec:full_pose_corr}
Translation based corrections rely on unnatural assumptions such as moving on a 3D plane only. A full pose correction is more general and can address a broader range of tasks. We evaluate this next. We compare DMPs with no adaptive terms (None / None), translation-only linear adaptations, similar to \cite{kumar2016optimal} (Linear / None), translation-only RL policy (PPO / None), similar to \cite{schoettler2019deep}. We also compare against different types of residual adaptive full-pose terms. We use the same linear policy from above and add Uniform random orientation corrections (Linear / Random), a full pose random policy (Random / Random), similar to \cite{pottle1963digital} and our full-pose RL policy (PPO / PPO).
\begin{table}[!htpb]
\Huge
\caption{\small Utilising full pose accuracy comparison. Pose baselines are named as: translation policy / orientation policy.}
\vspace{-2mm}
\begin{center}
\resizebox{0.49\textwidth}{!}{
\begin{tabular}{|c|c|c||c||c|c|c|}
\hline
\rbno{Type} & Corrections & \rano{Adaptive Policy} & Peg & Gear & RJ-45 & Average \\
\hline
\hline
A & \dmp{No corrections} & None / None & 52.6\% \huge{$\pm$ 0.7} & 41.5\% \huge{$\pm$ 1.5} & 0.0\% \huge{$\pm$ 0.0} & 31.4\% \huge{$\pm$ 0.7}\\
\hline
A & \dmp{translation} & Linear / None & 80.7\% \huge{$\pm$ 4.0} & 58.7\% \huge{$\pm$ 1.1} & 14.6\% \huge{$\pm$ 1.6} & 51.3\% \huge{$\pm$ 2.2}\\
\hline
D & \task{translation} & PPO / None & 94.2\% \huge{$\pm$0.9} & 50.0\% \huge{$\pm$0.4} & 57.2\% \huge{$\pm$2.1} & 67.4\% \huge{$\pm$1.1}\\
\hline
A & \dmp{full pose} & Linear / Random (ours) & 60.3\% \huge{$\pm$ 2.9} & 76.2\% \huge{$\pm$ 2.6} & 57.3\% \huge{$\pm$ 2.5} & 64.6\% \huge{$\pm$ 2.7}\\
\hline
A & \dmp{full pose} & Random / Random (ours) & 91.0\% \huge{$\pm$1.9} & 86.9\% \huge{$\pm$1.7} & 64.8\% \huge{$\pm$1.2} & 80.9\% \huge{$\pm$1.6}\\
\hline
D & \textbf{\task{full pose}} & \textbf{PPO / PPO (ours)} & \textbf{97.9\% \huge{$\pm$ 1.2}} & \textbf{92.2\% \huge{$\pm$ 2.6}} & \textbf{70.6\% \huge{$\pm$ 1.4}} & \textbf{86.9\% \huge{$\pm$ 1.7}}\\
\hline
\end{tabular}}
\end{center}
\label{tabl:physical_accuracy}
\vspace{-2mm}
\end{table}
Table~\ref{tabl:physical_accuracy} summarises our results. Full-pose rLfD performs best. A translation-only RL policy can be useful for simpler tasks, such as the round peg, but not when orientation matters. Random can harm performance on translation-only tasks (such as peg), but it can be useful for tasks like Gear, which heavily depends on accurate orientations. Random noise may result in better accuracy but it requires larger magnitude actions which damages fragile tips like RJ-45 so it should be preferably avoided.

\paragraph{Utilising full-pose nonlinear policies}
Linear and random solutions are susceptible to latent external factors in the environment, reducing performance. Compared to the perfectly levelled simulated surface, our physical set-up has a $1^{\circ}$ slope of the surface a socket is positioned on. Such changes may be hard to notice, but coping with them is not always straightforward and is therefore important.
Figure~\ref{fig:pos} disentangles the results from Table~\ref{tabl:physical_accuracy}. It can be seen that RL significantly increases accuracy on out-of-distribution start configurations \ra{($\pm3$cm and $\geq 20^{\circ}$)} across all three tasks. This indicates that an RL policy can be more effective at coping with latent external variability factors. \rb{This is likely due to the better generalisation of the full-pose RL corrections when compared to translation-only, analytical or non-residual solutions. While pure RL solutions may perform just as well, training them is challenging in real-world settings and often impractical, as discussed in Section~\ref{sec:applying_corrections}.d As a result, we could not extract a successful pure RL policy on the physical robot. More details are available on the website.}
\paragraph{Speed of Execution}
The final speed of insertion is another important factor. Ideally, a successful policy will be highly accurate, safe and fast. We report our findings in Table~\ref{tabl:speed_of_insertion}. The nonlinear adaptation policy was the fastest.
\begin{table}[!htpb]
\Huge
\caption{\small Achieved speed of insertion comparison. Pose baselines are named as: translation policy / orientation policy.}
\vspace{-2mm}
\begin{center}
\resizebox{0.48\textwidth}{!}{
\begin{tabular}{|c|c|c|c|c|c|c|}
\hline
\rbno{Type} & Corrections & \rano{Adaptive Policy} & Peg & Gear & RJ-45 & Average \\
\hline
\hline
A & \dmp{No corrections} & None / None & 7.0sec \huge{$\pm$0.1} & 8.1sec \huge{$\pm$0.2} & 10sec \huge{$\pm$0.0} & 8.6sec \huge{$\pm$0.1}\\
\hline
A & \dmp{full pose} & Linear / Random (ours) & 7.1sec \huge{$\pm$0.2} & 6.6sec \huge{$\pm$0.1} & 8.7sec \huge{$\pm$0.1} & 7.5sec \huge{$\pm$0.1}\\
\hline
A & \dmp{full pose} & Random / Random (ours) & 6.3sec \huge{$\pm$0.1} & 6.2sec \huge{$\pm$0.1} & 8.6sec \huge{$\pm$0.1} & 7.0sec \huge{$\pm$0.1}\\
\hline
D & \textbf{\task{full pose}} & \textbf{PPO / PPO (ours)} & \textbf{5.1sec \huge{$\pm$0.1}} & \textbf{5.9sec \huge{$\pm$0.1}} & \textbf{8.4sec \huge{$\pm$0.0}} & \textbf{6.5sec \huge{$\pm$0.1}}\\
\hline
\end{tabular}}
\end{center}
\label{tabl:speed_of_insertion}
\vspace{-4mm}
\end{table}
rLfD significantly improves the performance of a base DMP policy and generalisation to out of distribution start configurations. 
\paragraph{\rb{Transferring Residual Policies across Tasks}}
A key benefit of rLfD is that it allows residual skill transfer in a few update steps, thanks to its jiggling exploration in task space. Next, we evaluate the performance of a policy trained on a source task $src$ (either Gear or RJ-45) and then transfer to an associated target task $targ$ (either RJ-45 or Gear).
\begin{table}[!htpb]
\Huge
\caption{\small Successful insertions on transfer, comparison. Pose baselines are named as: translation policy / orientation policy.}
\vspace{-2mm}
\begin{center}
\resizebox{0.48\textwidth}{!}{
\begin{tabular}{|c|c|c|c|c|c|c|}
\hline
\rbno{Type} & Corrections & \rano{Adaptive Policy} & Gear & RJ-45 & Average & Eff.\\
\hline
D & \task{full pose} & (full training) $\pi_{targ}$ & 92.2\% \huge{$\pm$ 2.6} & 70.6\% \huge{$\pm$ 1.4} & 81.4\% \huge{$\pm$ 2.0} & 500\\
\hline
D & \task{full pose} & (full training) $\pi_{src}$ & 85.4\% \huge{$\pm$ 1.4} & 54.5\% \huge{$\pm$ 3.2} & 69.9\% \huge{$\pm$ 2.3} & 500\\
\hline
D & \task{full pose} & (3-shot) $\pi_{targ}$ & 70.3\% \huge{$\pm$ 4.0} & 59.1\% \huge{$\pm$ 3.1} & 64.7\% \huge{$\pm$ 3.6} & 60\\
\hline
D & \task{full pose} &  (3-shot) $\pi_{src \rightarrow targ}$ & 92.0\% \huge{$\pm$ 2.1} & 70.6\%  \huge{$\pm$ 1.7} & 81.3\% \huge{$\pm 1.9$} & 60\\
\hline
\end{tabular}}
\end{center}
\label{tabl:transfer}
\vspace{-4mm}
\end{table}
Table~\ref{tabl:transfer} shows the performance of the transferred policies. Columns 4 5 refer to the target tasks. We use one demo for $\pi_b$ and transfer residual policies trained on the $src$ tasks to $targ$ using three update steps (or 60 episodes). This equates to $\sim$15 minutes of training, including resets. We denote transferred policy as $\pi_{src \rightarrow targ}$ and policies trained only on the $targ$ or $src$ tasks as $\pi_{targ}$ and $\pi_{src}$. We consider $\pi_{targ}$ and $\pi_{src}$ trained with the full budget of 500 episodes (or $\sim$2 hours) and also training $\pi_{targ}$ from scratch for 3 update steps only (60 episodes). Results show that rLfD allows for successful policy transfer on both tasks, requiring eight times less training.
\vspace{-2mm}
\section{CONCLUSIONS}
This work explores DMP adaptation for contact-rich insertion tasks. Results show that residual learning from demonstration (rLfD) using RL adaptive policies in task space improves the generalisation abilities of DMPs both in simulation and real-world experiments.
Results show that rLfD with full pose corrections is highly effective and produces a gentle to the joints solution that can transfer across tasks. 
Finally, rLfD with nonlinear policies was shown to find better solutions when compared to linear and random policies. Future extensions to this work include consideration of difficulty in terms of sequences of contacts, optimising the parameterisation of the solution and investigating ways to reduce the overall forces applied during insertion. We also envision using rLfD to other contact-rich manipulation tasks, such as painting \cite{huang2020toward}, for visual policy inputs and learning generalised across skills policies.
\vspace{-4mm}
\section*{ACKNOWLEDGMENT}
\vspace{-1mm}
The authors would like to thank Giovanni Sutanto, Ning Ye, Daniel Angelov, Martin Asenov and Michael Mistry for the valuable insights and discussions on drafts of this work.

\bibliographystyle{IEEEtran} 
\bibliography{IEEEexample}

\end{document}